\documentclass[conference]{IEEEtran}
\IEEEoverridecommandlockouts
\usepackage{cite}
\usepackage{amsmath,amssymb,amsfonts}
\usepackage{algorithmic}
\usepackage{graphicx}
\usepackage{textcomp}
\usepackage{xcolor}
\usepackage{multirow}
\usepackage{tabularx}
\usepackage{booktabs} 
\usepackage{array}
\usepackage{booktabs}
\usepackage{multirow}
\usepackage{siunitx}      
\usepackage[compatibility=false]{caption}
\usepackage{hyperref}
\usepackage[a4paper,margin=1in]{geometry}

\def\BibTeX{{\rm B\kern-.05em{\sc i\kern-.025em b}\kern-.08em
    T\kern-.1667em\lower.7ex\hbox{E}\kern-.125emX}}
\begin{document}

\title{\includegraphics[height=1em]{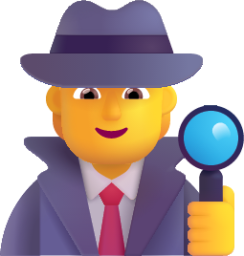} CoLMbo: Speaker Language Model for Descriptive Profiling\\
}

\author{\IEEEauthorblockN{Massa Baali$^{1*}$, Shuo Han$^{1*}$, Syed Abdul Hannan$^{1}$, Purusottam Samal $^{1}$, Karanveer Singh$^{2}$, \\Soham Deshmukh$^{1}$, Rita Singh$^{1}$, Bhiksha Raj$^{1}$}
\IEEEauthorblockA{
$^{1}$ Carnegie Mellon University, Pittsburgh, USA \\
$^{2}$FPrime AI, Pittsburgh, USA \\
mbaali@andrew.cmu.edu}
\thanks{$^*$Equal Contribution}
}

\maketitle

\begin{abstract}
Speaker recognition systems are often limited to classification tasks and struggle to generate detailed speaker characteristics or provide context-rich descriptions. These models primarily extract embeddings for speaker identification but fail to capture demographic attributes such as dialect, gender, and age in a structured manner. This paper introduces CoLMbo, a Speaker Language Model (SLM) that addresses these limitations by integrating a speaker encoder with prompt-based conditioning. This allows for the creation of detailed captions based on speaker embeddings. CoLMbo utilizes user-defined prompts to adapt dynamically to new speaker characteristics and provides customized descriptions, including regional dialect variations and age-related traits. This innovative approach not only enhances traditional speaker profiling but also excels in zero-shot scenarios across diverse datasets, marking a significant advancement in the field of speaker recognition. The code is available at: \url{https://github.com/massabaali7/CoLMbo}


\end{abstract}



\section{Introduction}
Speaker profiling is a critical challenge in speech processing, generating descriptive characterizations of speakers with applications in security, forensics, customer service, healthcare, and user interfaces \cite{singh2019profiling}. Traditional approaches \cite{r2024segaaunifiedapproachpredicting, Ravishankar2020PredictionOA} typically focus on predicting specific parameters such as age, gender, and accent using dedicated models. This method, however, is inflexible; it limits predictions to pre-defined parameters and requires new models for additional traits, without offering rich narrative descriptions.

Recent advances in audio analysis have explored the use of language models, but their application to speaker profiling remains underexplored. Existing Audio Language Models (ALMs), such as SaLMon \cite{tangsalmonn}, Qwen Audio \cite{chu2023qwen}, Pengi \cite{deshmukh2023pengi}, LTU \cite{gong2023listen}, and Gama \cite{ghosh2024gama}, primarily focus on general audio descriptions or phonetic content. These models often overlook detailed speaker characteristics, such as dialect, gender, or age, limiting their utility for comprehensive speaker profiling.

A key requirement in speaker profiling is ensuring consistency across multiple recordings of the same speaker. Attributes derived from different recordings of an individual should be identical, and their embedding-level representations should be highly similar, yet distinct from those of other speakers, even if their profiles share similar attributes. This consistency is critical for reliable profiling but is not adequately addressed by existing ALMs, which often fail to maintain speaker-specific distinctions in embeddings.

To address these challenges, we present CoLMbo, a unified Speaker Language Model (SLM) designed to generate detailed, natural language speaker profiles. By leveraging prompt-based conditioning, CoLMbo predicts individual attributes (e.g., gender, age) and produces comprehensive descriptions that integrate multiple characteristics. New attributes can be incorporated simply by augmenting the training data, without modifying the model architecture. To our knowledge, CoLMbo is the first SLM to offer such dynamic and descriptive speaker profiling capabilities.

CoLMbo employs a prefix-based architecture, where a sequential prefix derived from audio is fed into a text-based language model to generate descriptions. An initial token in the prefix is optimized to promote speaker-specific clustering in the embedding space. In addition to the standard cross-entropy loss for text generation, we include a speaker clustering loss to enhance consistency. Experiments show that this approach yields highly accurate text descriptions and near-perfect predictions of profile attributes.

CoLMbo achieves state-of-the-art performance in generating contextually rich speaker descriptions and predicting attributes like gender, age, accent, and ethnicity with near-perfect accuracy. These attributes serve as familiar benchmarks for evaluation, though CoLMbo’s flexible design supports additional characteristics through appropriate training data. This comprehensive approach marks a significant advancement in speaker profiling, expanding possibilities for both research and real-world applications.

\section{Literature Review}
Despite significant advancements in speaker recognition, limited attention has been devoted to developing robust Speaker Language Models capable of reliably capturing demographic characteristics such as dialect, gender, and age. Traditional speaker verification and recognition systems primarily focus on extracting speaker embeddings based on vocal features, often neglecting broader sociolinguistic and demographic factors that could enhance speaker profiling \cite{snyder2019speaker, dehak2011front, variani2014deep, wan2018generalized}. Recent research has highlighted the importance of demographic metadata in improving speaker recognition accuracy. Conventional approaches typically operate in a text-independent manner, relying solely on voice patterns while disregarding critical speaker-related metadata such as age, gender, and dialect \cite{dehak2010front}. While embedding-based speaker verification models \cite{gulati2020conformer, zhang2022mfa, lei2014novel, rahman2018employing, liu2018speaker} have significantly improved automatic speaker verification (ASV), they remain limited in capturing higher-level speaker attributes that extend beyond identity recognition. Several studies have attempted to address this gap by incorporating gender and age classification into speaker recognition. For example, \cite{kwasny2021gender} investigated deep neural networks (DNNs) such as x-vector and d-vector for automatic gender classification and age estimation. However, these models remain inherently classification-based, reducing speakers to discrete categories rather than generating rich, descriptive profiles. Similarly, deep learning approaches for metadata prediction continue to struggle with holistic speaker characterization, often failing to generalize beyond pre-trained categories and capturing dialectal variations and demographic traits in a meaningful way \cite{r2024segaaunifiedapproachpredicting, Ravishankar2020PredictionOA}. 
One of the most recent and relevant works \cite{wu2024explainable} employs a Concept-Bottleneck Model (CBM) to enhance interpretability by classifying speaker attributes such as gender, dialect, and age. While CBMs improve transparency, they are limited by their reliance on predefined classifiers, restricting their generalizability to unseen speaker traits and requiring labeled training data. Additionally, CBMs lack generative capabilities, as they can only classify attributes rather than generate descriptive speaker profiles. In contrast, CoLMbo leverages a generative speaker language model, enabling contextualized speaker descriptions without the need for explicit attribute classifiers or retraining, making it more flexible and scalable for diverse and zero-shot speaker profiling.

\section{CoLMbo architecture}
\begin{figure}[t]
  \centering
  \includegraphics[width=0.49\textwidth]{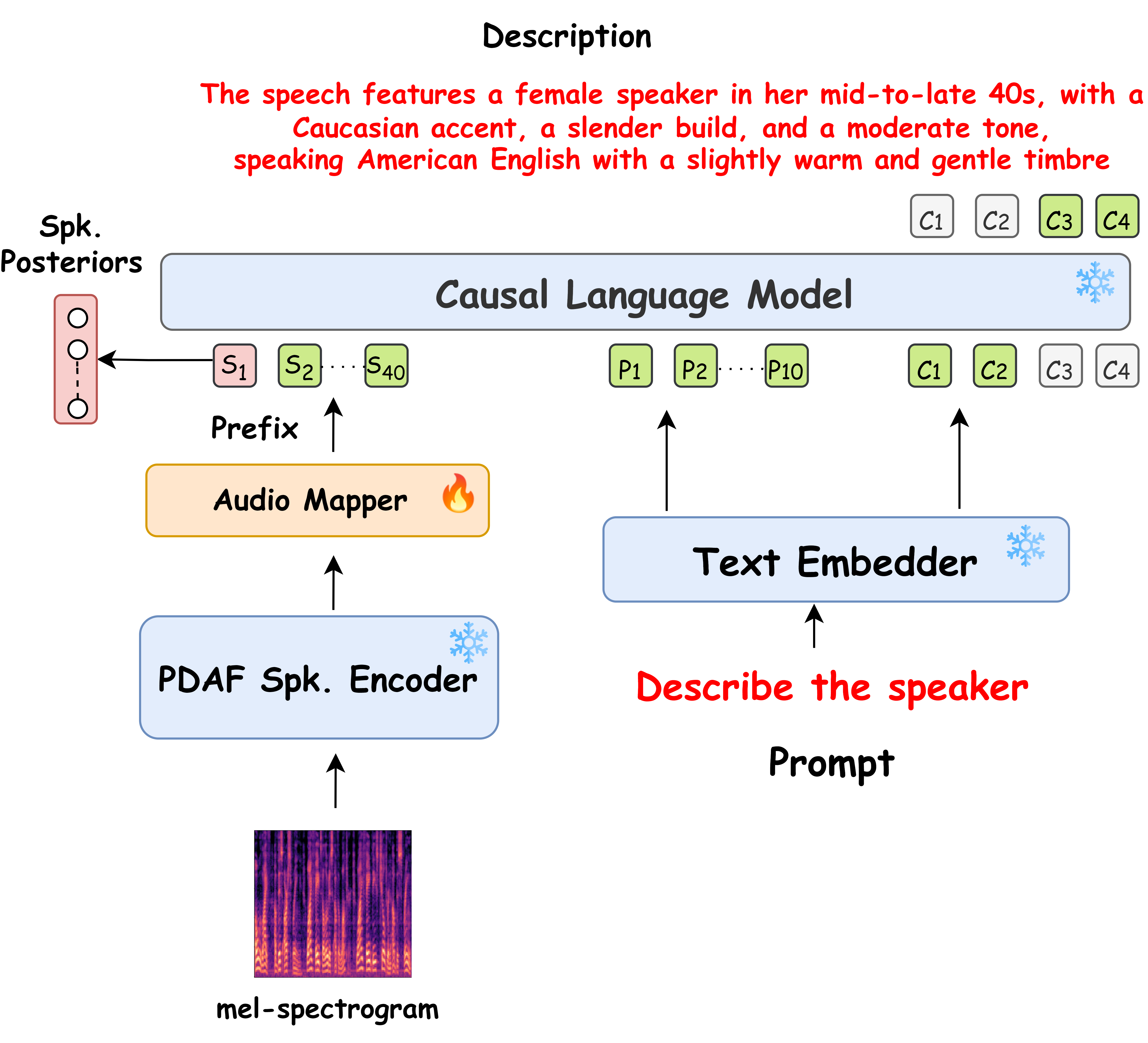}
  \caption{CoLMbo proposed Speaker Language Model.}
  \label{fig:CoLMboArchi}
\end{figure}

Figure \ref{fig:CoLMboArchi} depicts the architecture of the CoLMbo system. The modules of the model are described below.
\subsection{Audio parameterization} Input audio are first parameterized into a Mel-frequency spectro-
gram. The spectrogram is input into a speaker encoder which derives
a speaker embedding from it. In our work, we use the PDAF encoder
\cite{baalipdaf}, which extracts a 1024-dimensional speech embedding vector
from each recording. The speaker encoder remains fixed and is not updated during training.

\subsection{Audio Mapper} 
The speech embedding is then used to create an audio prefix to be input to the causal language model using a \textit{trainable} lightweight transformer-based mapping network. This mapping network \cite{mokady2021clipcap}, denoted as \(M\), takes the PDAF embedding as input and maps it to a sequence of \(k\) embeddings, as defined in equation \ref{eq:map}
\begin{equation}
p_1^i, \ldots, p_k^i = M(\text{PDAF}(x^i)).
\label{eq:map}
\end{equation}

Our implementation of the mapper follows that of  \cite{mokady2021clipcap}. It includes an initial fully-connected $1024 \times 30720$ layer followed by a ReLU activation, which produces a 30720-dimensional vector. This vector is reshaped into a $40 \times 768$ matrix, which is processed by 8 transformer layers, resulting in the final sequence of 40 768-dimensional audio prefix vectors.

Alternatively, we experiment with a Multi-Layer Perceptron (MLP) mapper composed of stacked fully-connected layers. The MLP takes as input the speech embedding and applies a series of linear transformations interleaved with nonlinear activations (e.g., Tanh), ultimately projecting the embedding to a 30,720-dimensional vector. This vector is then reshaped into a $40 \times 768$ matrix, corresponding to a sequence of 40 audio prefix tokens. Compared to the transformer-based mapper, this feedforward architecture is significantly lighter, while still enabling prefix conditioning through learned projections.



\subsection{Causal Language Model}
To generate text output, we utilize a pre-trained autoregressive causal language model, which remains frozen during both training and inference \cite{tsimpoukelli2021multimodal}. The audio prefix sequence generated by the mapper is combined with a text prompt prefix which could be a question or any other form of natural language prompt to form a final fixed-length prefix. This final prefix is used as input to the causal language model. Although the language model itself was not updated, it does pass gradients through to the mapping network \( M \), allowing it to be optimized. During inference, the language model generates tokens autoregressively, conditioned on both the speech and text prefixes.


\subsection{Speaker Classifier}
The first audio prefix vector is also passed to a speaker classifier, which classifies the speaker. The ``vocabulary'' of this classifier is the set of speakers in the training set.  Since the classifier is linear, it forces the first prefix vector to cluster by speaker, ensuring that the pattern of audio prefix vectors retains information about the speaker.

\subsection{Training}
%
CoLMbo is trained on a next-token prediction task, where tokens are generated autoregressively, conditioned on the prefix, to predict the descriptions 
$d^{i} = {\{d^{i}_1,...,d^{i}_l}\}$ conditioned on a prefix constructed from the audio input and text prompt. We use standard captioning loss as the primary loss function ($\mathcal{L}_1$) and the Cross-Entropy loss ($\mathcal{L}_2$) for speaker-id as the secondary loss function.
\begin{equation} 
\mathcal{L} = \alpha \mathcal{L}_1 +(1- \alpha)\mathcal{L}_2 \label{eq:loss} 
\end{equation}
\begin{equation}
    \mathcal{L}_1 = - \sum_{i=1}^{N} \sum_{j=1}^{l} \log p_\gamma(d_{ij} \mid p_1^i, \dots, p_{2k}^i, d_1^i, \dots, d_{j-1}^i) 
\end{equation}
\begin{equation}
    \mathcal{L}_2(y, \hat{y}) = - \sum_{i=1}^{C} y_i \log(\hat{y}_i)
\end{equation}
where N represents the number of text tokens and C is a number of speaker classes. y is a one-hot vector representing the true class and $\hat{y}$ is a vector of predicted probabilities.
\begin{table}[t]
\centering
\caption{Caption and Audio Statistics across Splits for TEARS Datasets}
\label{tab:tears-summary}
\resizebox{1.0\linewidth}{!}{%
\begin{tabular}{l|ccc|ccc}
\toprule
\textbf{Split} & \textbf{Vocab} & \textbf{Median} & \textbf{Max} & 
\textbf{Samples} & \textbf{Spks} & \textbf{Avg Duration} \\
\midrule
\textbf{Train} & 2334 & 51.0 & 120 & 26,126 & 562 & 8.72 $\pm$ 2.66  \\
\midrule
\textbf{Val} & 1071 & 51.0 & 85 & 5,424 & 100 & 9.89 $\pm$ 0.52 \\
\midrule
\textbf{Test} & 902 & 50.0 & 142 & 39,515 & 292 & 7.82 $\pm$ 5.23 \\
\bottomrule
\end{tabular}%
}
\end{table}


    \section{TEARS dataset}
To create (audio-text)-text triplets, comprising{speaker audio, prompt} and response, we utilize both the Expressive Anechoic Recordings of Speech (EARS) dataset \cite{richter2024ears} and the TIMIT Acoustic-Phonetic Continuous Speech Corpus \cite{garofolo1993darpa}. The EARS dataset provides a diverse range of speaking styles, including emotional speech, different reading styles, non-verbal sounds, and spontaneous conversational speech, along with speaker metadata such as gender and age, making it well-suited for speaker profiling. The TIMIT dataset, on the other hand, offers phonetically diverse speech samples from speakers across multiple dialect regions of American English, ensuring robust linguistic variability. We leverage metadata from both datasets to prompt an LLM (Llama3) \cite{touvron2023llama} to generate natural language descriptions of speakers, following established prompting strategies \cite{gong2023listen,deshmukh2024audio}. These descriptions capture speaker characteristics in a dynamic, free-form manner, extending beyond categorical labels. To further enhance granularity, we generate additional (audio-text)-text triplets focusing on gender, dialect, and ethnicity, integrating both datasets to create a comprehensive speaker profiling resource. We refer to this expanded (audio-text)-text dataset as TEARS. The caption and audio statistics for our dataset are presented in Table \ref{tab:tears-summary}, across training, validation, and test splits, and a higher vocabulary diversity in the training set to support robust generalization.

\begin{table*}[h!]
\centering
\caption{Performance of Speaker Profiling Models across datasets: Displays accuracy (Acc(\%)) and BERT (BT) scores for various speaker attributes, including age, gender, ethnicity, and dialect across different models. ``--'' indicates data unavailable.}
\resizebox{0.9\textwidth}{!}{%
\begin{tabularx}{\linewidth}{@{}lcc *{6}{>{\centering\arraybackslash}X}@{}}
    \toprule
    \textbf{Model} & \textbf{Params} & \textbf{Task} & \multicolumn{2}{c}{\textbf{VoxCeleb1-O}} & \multicolumn{2}{c}{\textbf{EARS}} & \multicolumn{2}{c}{\textbf{TIMIT}}\\
    \cmidrule(lr){4-5} \cmidrule(l){6-7} \cmidrule(l){8-9}
    & & & \textbf{Acc} & \textbf{BT} & \textbf{Acc} & \textbf{BT} & \textbf{Acc} & \textbf{BT}  \\
    \midrule
    \multirow{4}{*}{Qwen Audio} & \multirow{4}{*}{7.7B} & Age & 30.89 & 0.88 & 73.54 & 0.55 & 71.40 & 0.80 \\
    & & Gender & 61.66 & 0.93 & 51.00 & 0.72 & 71.31 & 0.93 \\
    & & Ethnicity & 59.59 & 0.78 & -- & -- & 7.60 & 0.79 \\
    & & Dialect & -- & -- & 26.93 & 0.61 & -- & -- \\
    \addlinespace
    \multirow{4}{*}{CoLMbo} & \multirow{4}{*}{24.85M} & Age & 91.80 & 0.97 & 93.98 & 0.87 & 97.87 & 0.87 \\
    & & Gender & 82.75 & 0.90 & 85.90 & 0.90 & 83.37 & 0.90 \\
    & & Ethnicity & 88.38 & 0.96 & -- & -- & 86.42 & 0.97 \\
    & & Dialect & -- & -- & 88.58 & 0.98 & -- & -- \\
    \addlinespace
    \bottomrule
    
\end{tabularx} }
\label{tab:results}
\end{table*}

\section{Experimental Setup} \label{sec:exp}
\begin{figure*}[htb]
	
		\centering
		\centerline{\includegraphics[width=15.5cm]{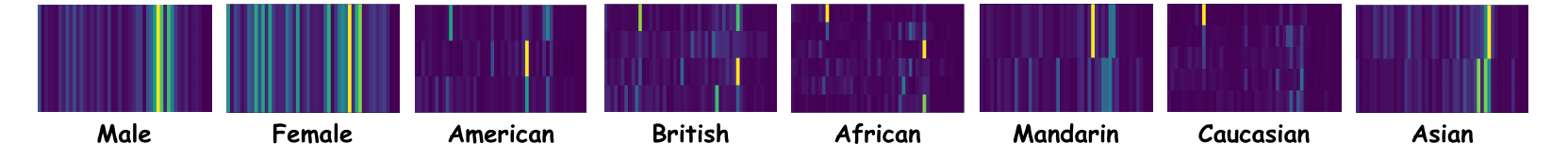}}
	\caption{CoLMbo Attention maps showing distinct activation patterns for speaker attributes}
	\label{fig:CoLMboattnMap}
\end{figure*}
The audio data from EARS was resampled to 16 kHz and split into 26,000 training, 600 validation, and 1,600 test utterances. TIMIT was used without resampling, comprising 4,620 training and 1,680 test utterances. To enhance model robustness, we applied the following audio augmentations dynamically during training: (1) Add Reverberation, using a room impulse response with decay times of 0.5–2 seconds; (2) Add Noise, introducing white noise at 10–20 dB SNR; (3) Drop Frequency, attenuating random frequency bands (e.g., 500–2000 Hz); and (4) Random Time Cut, removing 100–500 ms segments. The augmented raw waveforms were converted into 128-dimensional Log-Mel Spectrograms using a 25 ms Hamming window and a 10 ms stride, and served as inputs to the speaker encoder. Prompts were tokenized using a BPE tokenizer, forming a fixed-length prefix of 50 tokens - 40 from the audio and 10 from the text prompt.

CoLMbo is trained end-to-end using the Adam optimizer \cite{kingma2015adam} with a learning rate of 1e-4, linear warmup, and batch size of 64. The model is first trained with the language model frozen for 100 epochs, followed by full fine-tuning of all components for an additional 30 epochs. This staged training regime stabilizes alignment between modalities before enabling full model adaptation. The training objective jointly optimizes a weighted combination of captioning loss and speaker classification loss, with $\alpha$ set to 0.3.

Full fine-tuning is feasible in our setup due to the use of a small language model (GPT2-base), which allows for efficient training without the computational demands of large-scale LLMs. All models were implemented in PyTorch \cite{paszke2019pytorch} and trained on eight NVIDIA V100 GPUs.
\begin{figure}[t]
  \centering
  \includegraphics[height=0.23\textheight]{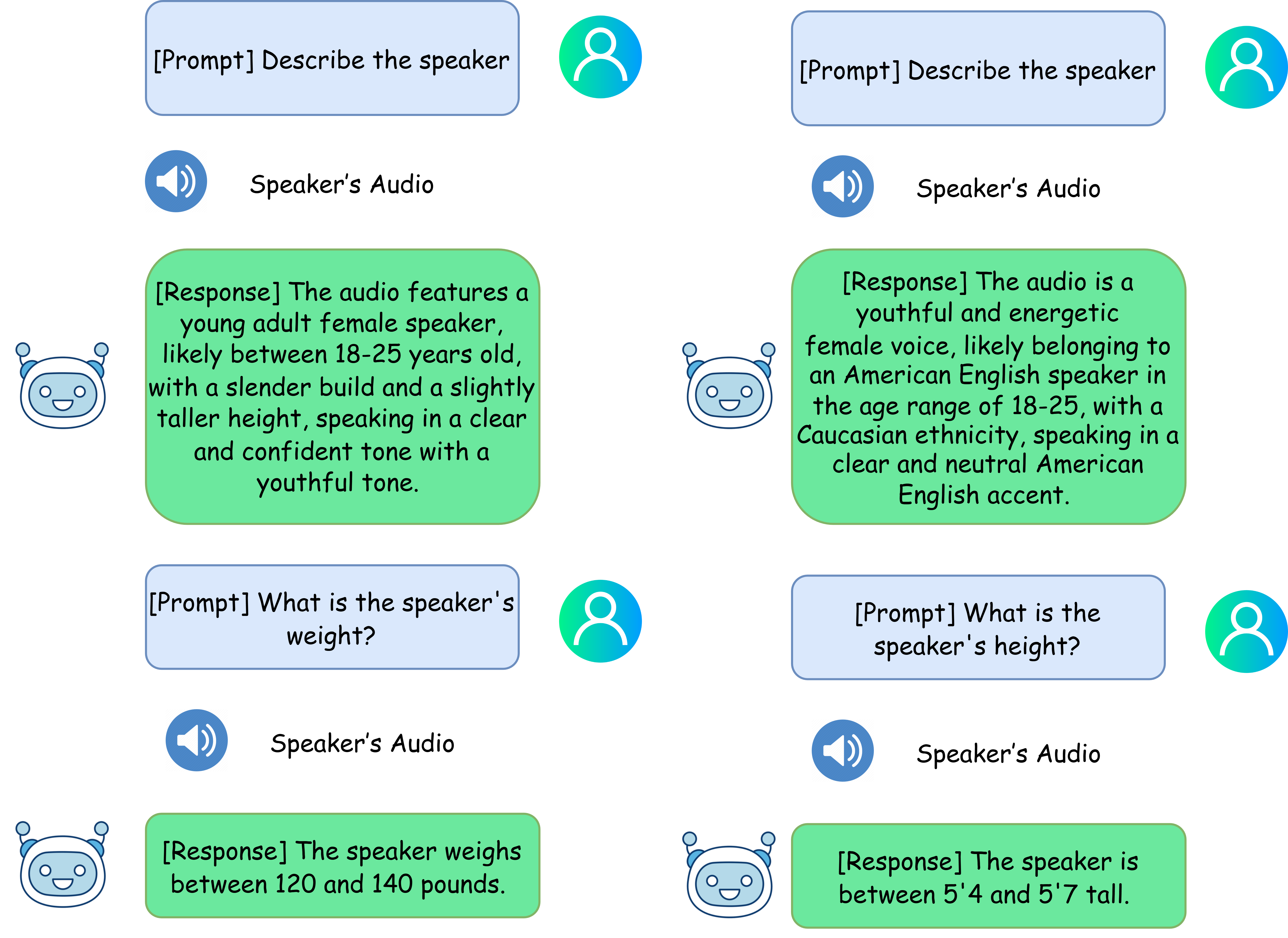}
  \caption{Illustrative examples of prompt‐response for CoLMbo. Each panel shows (1) user prompt, (2) speaker’s input audio and (3) the CoLMbo’s generated response in a green chat bubble }
  \label{fig:CoLMboPrompt}
\end{figure}

\section{Results and Analysis}
\begin{figure*}[t]
  \centering
  \includegraphics[width=0.85\textwidth]{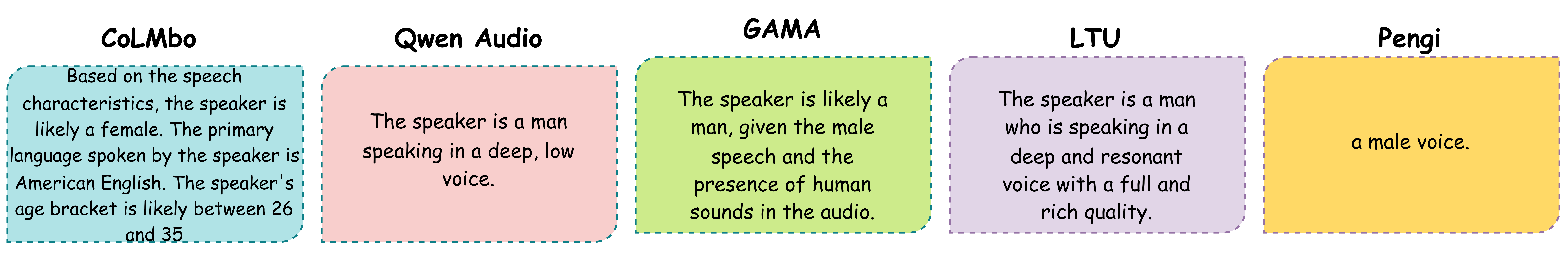}  

  \caption{Comparison of CoLMbo and Other ALMs in Generating Speaker Features on Unseen Speakers in a Zero-Shot Setting}
  \label{fig:alms}
\end{figure*}

In this section, we demonstrate the effectiveness of our system in generating detailed and descriptive speaker characteristics.
\subsection{Interpretability}
CoLMbo is capable of extracting meaningful insights from the sequence of prefix vectors, which can be inferred by looking at the attention maps of the audio mapper, offering an interpretable explanation of how the model focuses on different aspects of the speech. In the attention maps of the audio mapper illustrated in the figure \ref{fig:CoLMboattnMap}, which is used for generating speaker descriptions, we observe that specific characteristics such as gender, age, ethnicity, and accent trigger distinct sets of activation patterns across the prefix vectors fed into the language model. For instance, as indicated by the attention maps, characteristics like “Female,” “African,” and “Mandarin” result in a unique set of activations that emphasize different portions of the prefix sequence compared to “Male,” “Caucasian,” or “British.” These activations reveal how the model distinctly processes various characteristics through characteristic-specific prefixes. While the embeddings themselves remain constant, the attention map shows how the pattern of attention across different prefixes shifts to reflect variations in speaker attributes. This suggests that the model effectively discriminates between different values of speaker characteristics by altering the activation patterns, thus allowing for the generation of nuanced and accurate descriptions.
This behavior highlights the importance of the prefix vectors in encoding different speaker characteristics and how they are dynamically activated based on the input speaker’s attributes. As a result, different characteristics, while rooted in similar embeddings, only differ in how they selectively activate certain parts of the prefix, leading to characteristic-specific descriptive outputs.
\begin{table*}[ht!]
\centering
\caption{Ablation results for CoLMbo: (Acc(\%)) and BERTScore (BT) across datasets. ``--'' indicates unavailable data.}
\label{tab:ablation}
\resizebox{0.8\textwidth}{!}{%
\begin{tabularx}{\linewidth}{@{}l l l l *{6}{>{\centering\arraybackslash}X}@{}}
    \toprule
    \textbf{Mapper} & \textbf{Speaker Loss} & \textbf{Finetune LM} & \textbf{Task} & \multicolumn{2}{c}{\textbf{VoxCeleb2}} & \multicolumn{2}{c}{\textbf{EARS}} & \multicolumn{2}{c}{\textbf{TIMIT}} \\
    \cmidrule(lr){5-6} \cmidrule(lr){7-8} \cmidrule(l){9-10}
    & & & & \textbf{Acc} & \textbf{BT} & \textbf{Acc} & \textbf{BT} & \textbf{Acc} & \textbf{BT} \\
    \midrule
    \multirow{4}{*}{MLP} & \multirow{4}{*}{\checkmark} & \multirow{4}{*}{\checkmark} & Age       & 91.80 & 0.97   & 93.98 & 0.87   & 97.87 & 0.87   \\
    & & & Gender    & 82.75 & 0.90   & 85.90 & 0.90   & 83.37 & 0.90   \\
    & & & Ethnicity & 88.38 & 0.96   & -- & --   & 86.42 & 0.97   \\
    & & & Dialect   & --    & --   & 88.58 & 0.98   & --   & --   \\
    \addlinespace
    \multirow{4}{*}{MLP} & \multirow{4}{*}{} & \multirow{4}{*}{\checkmark} & Age       & 90.20 & 0.97   & 93.88 & 0.86   & 96.92 & 0.87   \\
    & & & Gender    & 80.08 & 0.90   & 83.59 & 0.90   & 83.78 & 0.90   \\
    & & & Ethnicity & 86.71 & 0.95   & -- & --   & 77.78 & 0.97   \\
    & & & Dialect   & --    & --   & 90.02 & 0.98   & --    & --   \\
    \addlinespace
    \multirow{4}{*}{Transformer} & \multirow{4}{*}{\checkmark} & \multirow{4}{*}{\checkmark} & Age       & 91.67 & 0.97   & 94.02 & 0.87   & 97.98 & 0.89   \\
    & & & Gender    & 85.07 & 0.91   & 87.06 & 0.90   & 85.28 & 0.91   \\
    & & & Ethnicity & 86.15 & 0.96   & -- & --   & 78.38 & 0.97   \\
    & & & Dialect   & --    & --   & 87.21 & 0.98   & --    & --   \\
   \addlinespace
    \multirow{4}{*}{MLP} & \multirow{4}{*}{\checkmark} & \multirow{4}{*}{} & Age       & 91.07 & 0.97   & 92.96 & 0.86  & 98.90 & 0.87   \\
    & & & Gender    & 80.19 & 0.89   & 85.40 & 0.91   & 87.39 & 0.90   \\
    & & & Ethnicity & 87.45 & 0.95   & -- & --   & 100.00 & 0.97   \\
    & & & Dialect   & --    & --   & 89.09 & 0.98   & --    & --   \\
    \addlinespace
    \bottomrule
\end{tabularx}
}
\end{table*}
\begin{figure*}[t]
  \centering
  \includegraphics[width=0.78\textwidth]{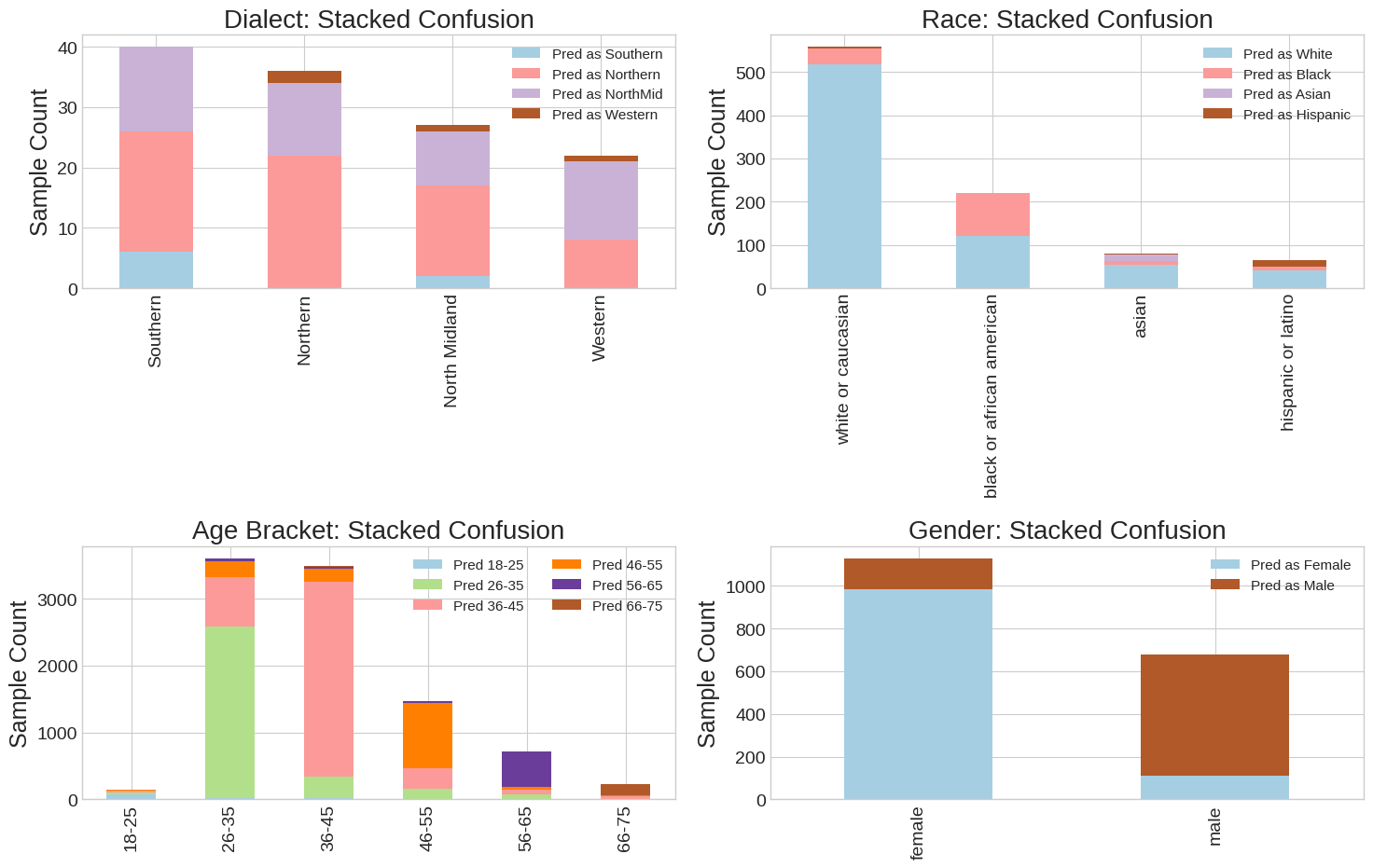}
  \caption{Stacked confusion plots for CoLMbo on TEARS, revealing bias in dialect and ethnicity predictions due to dataset imbalances, mitigated by audio augmentations.}
  \label{fig:stackConf}
\end{figure*}

\subsection{Speaker descriptions}
We conduct a comparative analysis of speaker profiling models, specifically CoLMbo and Qwen Audio, as shown in Table \ref{tab:results}. Despite having significantly fewer parameters (24.85M vs. 7.7B), CoLMbo consistently demonstrates superior performance across multiple attributes on VoxCeleb1-O. For instance, CoLMbo achieves 91.80\% accuracy in age prediction with a BERT score (BT) of 0.97, compared to Qwen Audio’s 30.89\% and 0.88, respectively. Similarly, CoLMbo attains 82.75\% accuracy in gender prediction and 88.38\% in ethnicity, surpassing Qwen Audio's 61.66\% and 59.59\% on the same tasks. Notably, CoLMbo also generalizes well across datasets. On the EARS dataset, it achieves 88.58\% accuracy and 0.90 BT for dialect prediction, outperforming Qwen Audio by a wide margin (26.03\%, 0.79 BT). On TIMIT, CoLMbo maintains high accuracy for age (97.90\%), gender (83.37\%), and ethnicity (86.42\%), while Qwen Audio lags behind across all metrics. These results underscore CoLMbo’s robustness and generalizability despite its compact size. Importantly, CoLMbo was evaluated in a zero-shot setting on all datasets, highlighting the adaptability and strength of its underlying architecture.
The following figure~\ref{fig:CoLMboPrompt} further illustrates qualitative examples of generated speaker descriptions, showing CoLMbo’s outputs for unseen test-set speakers, underscoring its ability to produce coherent, contextually accurate descriptions.

In Figure~\ref{fig:alms}, we extend our evaluation by comparing CoLMbo against several alternative ALMs. Evaluated under zero-shot conditions, CoLMbo consistently outperforms competitors, highlighting its superior capability in accurately capturing speaker characteristics and offering detailed interpretability.
\begin{figure}[t]
  \centering
  \includegraphics[width=0.24\textwidth]{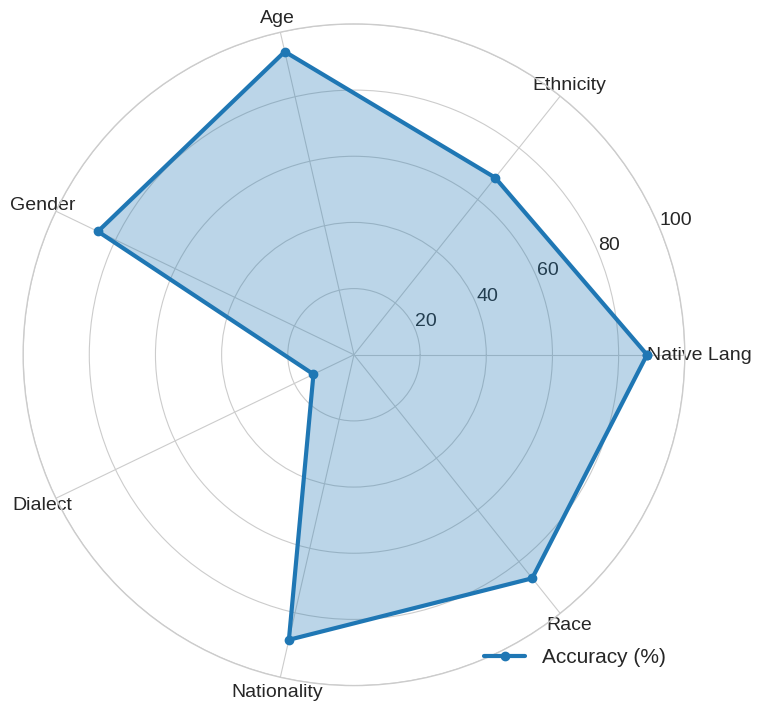}
  \caption{Radar plot of CoLMbo’s accuracy for speaker attributes.}
  \label{fig:radar}
\end{figure}

\subsection{Bias and Limitations}
While CoLMbo demonstrates strong performance, analysis of its behavior (Figures~\ref{fig:radar} and~\ref{fig:stackConf}) highlights opportunities to enhance its handling of certain attributes, particularly those affected by dataset imbalances. Figure~\ref{fig:radar} shows high accuracy for age, gender, nationality, and native language, with slightly lower performance on dialect and ethnicity, where subtle distinctions and limited training data present challenges common in speaker profiling. Figure~\ref{fig:stackConf} illustrates that for dialect, CoLMbo tends to predict more frequent classes like \textit{Northern} or \textit{North Midland,} sometimes misclassifying less-represented dialects like \textit{Southern} or \textit{Western.} This reflects the influence of dataset distribution and the complexity of capturing nuanced acoustic cues. Similarly, for race, predictions lean toward the majority class (\textit{White}), a common issue in voice-based models due to data skew. Age predictions tend to favor common age groups (26–35 and 36–45), sometimes assigning other age ranges to these ranges, consistent with data distribution effects. Gender prediction performs well overall, though minor misclassifications between male and female samples occur, likely due to some overlapping voice characteristics. These observations suggest that expanding training data diversity and refining acoustic feature extraction could further improve performance on underrepresented or ambiguous attributes.

\subsection{Ablations}
To better understand the components that contribute into CoLMbo's strong performance, we present an ablation study in Table~\ref{tab:ablation}. This table reports accuracy and BERTScore for age, gender, dialect and ethnicity prediction across multiple datasets. We systematically analyze variations in the architecture of the mapper, the inclusion of the auxiliary speaker loss, and the full fine-tuning of the language model.
We observe that:
\textbf{Speaker Loss:} Removing the speaker loss consistently reduces accuracy across most tasks and datasets. For instance, without this auxiliary objective, gender accuracy on VoxCeleb2 drops by over 2\% (from 82.75\% to 80.08\%), ethnicity accuracy on the Ethnicity dataset falls by more than 9\% (from 86.42\% to 77.78\%), and age estimation performance declines across all evaluated datasets. These results highlight that explicit speaker supervision enhances the model's ability to learn discriminative and generalizable representations.
\textbf{Audio Mapper:} MLP-based mapper performs as well as, or better than, the Transformer-based mapper. Notably, for the ethnicity attribute in the TIMIT dataset, the MLP variant slightly outperforms the Transformer, indicating that the simpler MLP architecture is sufficient for effective attribute mapping in this setting.
\textbf{LM Fine-Tuning:} Fine-tuning the language model consistently improves performance, particularly for challenging attributes like gender and ethnicity in zero-shot settings on VoxCeleb2. Fine-tuned LMs achieve more stable accuracy across datasets compared to frozen LMs. This approach is especially effective for generalizing to out-of-domain data, as evidenced by the enhanced results on VoxCeleb2.
\textbf{Audio Augmentations:} To evaluate the contribution of audio augmentations (Add Reverb, Add Noise, Drop Frequency, Random Time Cut) introduced in Section \ref{sec:exp}, we compared CoLMbo’s performance on the VoxCeleb1-O dataset with and without augmentations. Without augmentations, CoLMbo achieved an age prediction accuracy of 59.23\% and a gender prediction accuracy of 81.56\%. After applying augmentations, age accuracy improved significantly to 91.80\%, and gender accuracy slightly increased to 82.75\%. These improvements, particularly in age prediction, demonstrate that augmentations enhance CoLMbo’s robustness to diverse acoustic conditions, such as noise and reverberation, enabling better generalization in zero-shot settings. This underscores the importance of our augmentation strategy for achieving state-of-the-art performance in speaker profiling.

\section{Conclusion}
In this paper, we present CoLMbo, a unified SLM designed to capture demographic
attributes addressing the limitation of speaker recognition system by integrating a lightweight speaker encoder with prompt-based conditioning. CoLMbo generates detailed, context-aware speaker profiles that capture nuanced attributes, such as dialect, gender, and age. Our experiments demonstrate that CoLMbo consistently outperforms larger models across multiple datasets, achieving high accuracy in zero-shot speaker profiling without retraining. This robust performance, particularly in unseen conditions, highlights CoLMbo’s potential as a scalable and versatile solution for comprehensive speaker analysis.

In future work, we aim to enhance CoLMbo through improving the interpretability of its generated profiles to provide clearer insights into how attributes are inferred. To mitigate observed biases, such as the tendency to favor majority classes, we will collect more balanced datasets and apply debiasing techniques. Finally, we intend to expand CoLMbo’s capabilities by incorporating additional speaker characteristics, including emotional state, communication style, and health-related vocal markers, to enhance its adaptability and utility. 

\bibliographystyle{IEEEtran}
\bibliography{conference_101719}

\end{document}